# Probabilistic Structured Predictors

**Shankar Vembu, Thomas Gärtner, Mario Boley**
Fraunhofer IAIS
Schloß Birlinghoven
53754 Sankt Augustin, Germany
{shankar.vembu, thomas.gaertner, mario.boley}@iais.fraunhofer.de

## Abstract

We consider MAP estimators for structured prediction with exponential family models. In particular, we concentrate on the case that efficient algorithms for uniform sampling from the output space exist. We show that under this assumption (i) exact computation of the partition function remains a hard problem, and (ii) the partition function and the gradient of the log partition function can be approximated efficiently. Our main result is an approximation scheme for the partition function based on Markov Chain Monte Carlo theory. We also show that the efficient uniform sampling assumption holds in several application settings that are of importance in machine learning.

## 1 Introduction

We consider discriminative structured prediction models with an emphasis on predicting combinatorial structures. These structures find applications in machine learning problems such as multi-label classification (Elisseeff & Weston, 2001), multi-category hierarchical classification (Cesa-Bianchi et al., 2006), and label ranking (Dekel et al., 2003).

Let $\mathcal{X} \times \mathcal{Y}$ be the domain of observations and labels, and let $X = (x_1, \cdots, x_m) \in \mathcal{X}^m$, $Y = (y_1, \cdots, y_m) \in \mathcal{Y}^m$ be the set of observations. Our goal is to estimate $y|x$ using exponential families via

$$p(y|x, \theta) = \exp(\langle \phi(x,y), \theta \rangle - \ln Z(\theta|x)) \ ,$$

where $\phi(x, y)$ are the joint sufficient statistics of $x$ and $y$, and $Z(\theta|x) = \sum_{y \in \mathcal{Y}} \exp(\langle \phi(x,y), \theta \rangle)$ is the partition function which takes care of the normalisation. We perform maximum-a-posteriori (MAP) parameter estimation by imposing a normal prior on $\theta$. This leads to optimising the negative joint likelihood in $\theta$ and $Y$:

$$\hat{\theta} = \underset{\theta}{\operatorname{argmin}} \left[ -\ln p(\theta, Y|X) \right]$$
$$= \underset{\theta}{\operatorname{argmin}} \left[ \lambda \|\theta\|^2 + \frac{1}{m} \sum_{i=1}^{m} [\ln Z(\theta|x_i) - \langle \phi(x_i, y_i), \theta \rangle] \right] ,$$
(1)

where $\lambda > 0$ is the regularisation parameter. Throughout this paper, we assume that the $\ell_2$ norm of the sufficient statistics and the parameters are bounded, i.e., $\|\phi(x,y)\| \leq R$ and $\|\theta\| \leq B$, where $R$ and $B$ are constants (note that $B$ can be bounded from above as shown in Appendix C). The difficulty in solving (1) lies in the computation of the partition function. The optimisation is typically performed using gradient descent techniques (and advancements thereof). We therefore also need to compute the gradient of the log partition function, which is the first order moment of the sufficient statistics, i.e., $\nabla_\theta \ln Z(\theta|x) = \mathbb{E}_{y \sim p(y|x,\theta)}[\phi(x, y)]$.

Computing the log partition function and its gradient are in general NP-hard. In section 2, we will show that computing the partition function still remains NP-hard given a uniform sampler for $\mathcal{Y}$. We therefore need to resort to approximation techniques to compute these quantities. Unfortunately, application of concentration inequalities do not yield approximation guarantees with polynomial sample size. In this work, we present Markov chain Monte Carlo (MCMC) based approximations for computing the partition function and the gradient of the log partition function with provable guarantees. There has been a lot of work in applying Monte Carlo algorithms using Markov chain simulations to solve ♯P-complete counting and NP-hard optimisation problems. Recent developments include a set of mathematical tools for analysing the rates of convergence of Markov chains to equilibrium (see Randall (2003); Jerrum and Sinclair (1996) for surveys). To the best of our knowledge, these tools have not been applied in the design and analysis of structured prediction problems, but have been referred to as an important research frontier (Andrieu et al., 2003) for MCMC based machine learning problems in general.

### 1.1 Our Contributions

We analyse discriminative probabilistic structured prediction models based on exponential families using MCMC theory. In particular,

- we prove a hardness result for computing the partition function (Section 2);

- we present an algorithm for approximating the partition function and the gradient of the log partition function with provable guarantees (Section 2);

- we design a Markov chain that can be used to sample combinatorial structures from exponential family distributions considered in this work given that there exists an exact uniform sampler, and also perform a non-asymptotic analysis of its mixing time (Section 3);

- we describe several combinatorial structures that find applications in machine learning problems, including multi-label classification, label ranking, and multi-category hierarchical classification (Section 4).

**Notation:** We will use $[\![n]\!]$ to denote the set of integers $\{1, \ldots, n\}$, $||\cdot||$ to denote the $\ell_2$ norm of a vector, and $|\cdot|$ to denote the $\ell_1$ norm if the argument is a vector, the absolute value if the argument is a scalar, and the cardinality if the argument is a set.

## 2 Computing the Partition Function: Hardness Result and Approximations

### 2.1 Hardness of Computing the Partition Function

Consider the output space of undirected cycles over a fixed set of vertices $\Sigma$, i.e., $\mathcal{Y} = \bigcup_{U \subset \Sigma} \text{cyclic\_permutations}(U)$. Let $\psi : \mathcal{Y} \to \mathbb{R}^{\Sigma \times \Sigma}$ with $\psi_{uv}(y) = 1$ if $\{u, v\} \in y$ and 0 otherwise.

**Theorem 2.1** *Computing the partition function of cyclic permutations is NP-hard.*

**Proof** Suppose we can compute $\ln Z(\theta) = \ln \sum_{y \in \mathcal{Y}} \exp(\langle \psi(y), \theta \rangle)$ efficiently. Given an arbitrary graph $G = (V, E)$ with adjacency matrix $\bar{\theta}$, let $\Sigma = V$ and $\theta = \bar{\theta} \times \ln(|V|! \times |V|)$. We will show that $G$ has a Hamiltonian cycle if and only if $\ln Z(\theta) \geq |V| \times \ln(|V|! \times |V|)$.

First, observe that $|\mathcal{Y}| < |V|! \times |V|$.

Necessity: As the exponential function is positive and ln is monotone increasing, it follows that $\ln Z(\theta) \geq |V| \times \ln(|V|! \times |V|)$.

Sufficiency: Suppose G has no Hamiltonian cycle. Then

$$\ln Z(\theta) \leq \ln[|\mathcal{Y}| \times \exp[(|V| - 1) \times \ln(|V|! \times |V|)]]$$
$$= \ln |\mathcal{Y}| + (|V| - 1) \times \ln(|V|! \times |V|)$$
$$< |V| \times \ln(|V|! \times |V|) .$$

This completes the proof. □

We are interested in the class of problems for which sampling uniformly at random is easy, and cyclic permutations is one example of these. The following result shows that computing the partition function is hard even if we restrict the problem to this class. Essentially, it transfers the general NP-hardness result of computing the partition function to the restricted class of problems that we are interested in.

**Corollary 2.2** *Computing the partition function of cyclic permutations remains NP-hard even if efficient algorithms for uniform sampling of cyclic permutations exist.*

In the next section, we show how to approximate the partition function given that there exist efficient algorithms for uniform sampling.

### 2.2 Approximating the Partition Function

As a first step towards approximating the partition function, we could consider using concentration inequalities. If we can sample uniformly at random from $\mathcal{Y}$, then we can apply Hoeffding's inequality to bound the deviation of the partition function $Z(\theta|x)$ from its finite sample expectation $\hat{Z}(\theta|x)$. Unfortunately, the bound obtained from using Hoeffding's inequality is not useful due to its dependence on the size of the output space $|\mathcal{Y}|$. We now present an algorithm that is a *fully-polynomial randomised approximation scheme* for approximating the partition function.

**Definition 2.1** *Suppose $f : P \to \mathbb{R}^+$ is a function that maps problem instances $P$ to positive real numbers. A* randomised approximation scheme *for $P$ is a randomised algorithm that takes as input an instance $p \in P$ and an error parameter $\epsilon > 0$, and produces as output a number $Q$ such that*

$$\Pr[(1 - \epsilon)f(p) \leq Q \leq (1 + \epsilon)f(p)] \geq \frac{3}{4} .$$

*A randomised approximation scheme is said to be* fully polynomial (FPRAS) *if it runs in time polynomial in the length of $p$ and $1/\epsilon$.*

We exploit the intimate connection between counting and sampling problems (Jerrum et al., 1986) to approximately compute the partition function using sampling algorithms. The technique is based on a reduction from counting to sampling. The standard approach (Jerrum & Sinclair, 1996) is to express the quantity of interest, i.e., the partition function $Z(\theta|x)$, as a telescoping product of ratios of parameterised variants of the partition function. Let $0 = \beta_0 < \beta_1 \cdots < \beta_l = 1$ denote a sequence of parameters also called as *cooling schedule* and express $Z(\theta|x)$ as a telescoping product

$$\frac{Z(\theta|x)}{Z(\beta_{l-1}\theta|x)} \times \frac{Z(\beta_{l-1}\theta|x)}{Z(\beta_{l-2}\theta|x)} \times \cdots \frac{Z(\beta_1\theta|x)}{Z(\beta_0\theta|x)} \times Z(\beta_0\theta|x) \ .$$

Define the random variable $f_i(y) = \exp[(\beta_{i-1} - \beta_i)\langle\phi(x,y),\theta\rangle]$ (we omit the dependence on $x$ to keep the notation clear), for all $i \in [\![l]\!]$, where $y$ is chosen according to the distribution $\pi_{\beta_i} = p(y|x, \beta_i\theta)$. We then have

$$\mathbb{E}f_i = \sum_{y \in \mathcal{Y}} \exp[(\beta_{i-1} - \beta_i)\langle\phi(x,y),\theta\rangle] \frac{\exp[\beta_i \langle\phi(x,y),\theta\rangle]}{Z(\beta_i\theta|x)}$$
$$= \frac{Z(\beta_{i-1}\theta|x)}{Z(\beta_i\theta|x)} \ ,$$

which means that $f_i(y)$ is an unbiased estimator for the ratio $\rho_i = Z(\beta_{i-1}\theta|x)/Z(\beta_i\theta|x)$. This ratio can now be estimated by sampling according to the distribution $\pi_{\beta_i}$ and computing the sample mean of $f_i$. The desideratum is an upper bound on the variance of this estimator. Having a low variance implies a small number of samples $S$ to approximate each ratio. The final estimator is then the product of the reciprocal of the individual ratios in the telescoping product.

We now proceed with the derivation of an upper bound on the variance of the random variable $f_i$, or more precisely on the quantity $B_i = \mathrm{Var} f_i/(\mathbb{E}f_i)^2$. We first assume that $Z(\beta_0\theta|x) = |\mathcal{Y}|$ can be computed in polynomial time[1]. We use the following cooling schedule (Stefankovic et al., 2007): $l = p\lceil R||\theta||\rceil$; $\beta_j = j/(pR||\theta||)$ for all $j \in [\![l-1]\!]$, where $p$ is a constant integer $\geq 3$, i.e., we let the cooling schedule to be of the following form:

$$0, \frac{1}{q}, \frac{2}{q}, \frac{3}{q}, \cdots, \frac{p\lfloor R||\theta||\rfloor}{q}, 1 \ ,$$

where $q = pR||\theta||$ (and w.l.o.g. we assume that $R||\theta||$ is non-integer and real). Given this cooling schedule, observe that $\exp(-1/p) \leq f_i \leq \exp(1/p)$, which follows from the definition of the random variable $f_i$, and also $\exp(-1/p) \leq \mathbb{E}f_i = \rho_i \leq \exp(1/p)$. We are now ready to prove the bound on the quantity $B_i$.

---
[1] The assumption is true in all the applications we consider in this work. If it is not possible to compute $|\mathcal{Y}|$, which is a counting problem, in polynomial time, we can still approximate it using the machinery introduced in this section.

**Proposition 2.3** $B_i = \frac{\mathrm{Var} f_i}{(\mathbb{E}f_i)^2} \leq \exp(2/p), \ \forall \ i \in [\![l]\!]$.

We first need to prove the following lemma.

**Lemma 2.4** $\exp(1/p) - 1 \leq \rho_i \leq \exp(-1/p) + 1$.

**Proof** $a \leq b \Rightarrow \exp(a) - \exp(-a) \leq \exp(b) - \exp(-b)$ as the exponential function is monotone increasing. Thus $a \leq 1/3 \Rightarrow \exp(a) - \exp(-a) \leq \exp(1/3) - \exp(-1/3) < 1$. Setting $a = 1/p$ with $p \geq 3$ and using the fact that $\exp(-1/p) \leq \rho_i \leq \exp(1/p)$ for all $i \in [\![l]\!]$ proves the lemma. $\square$

**Proof** (of Proposition 2.3) Consider $\rho_i \geq \exp(1/p) - 1 \geq f_i - 1$. This implies $f_i - \rho_i \leq 1$. Next, consider $\rho_i \leq \exp(-1/p) + 1 \leq f_i + 1$. This implies $f_i - \rho_i \geq -1$. Combining these, we get $|f_i - \rho_i| \leq 1$, which implies $\mathrm{Var} f_i \leq 1$, and therefore $\mathrm{Var} f_i/(\mathbb{E}f_i)^2 \leq \exp(2/p)$. $\square$

Equipped with this bound, we are ready to design an FPRAS for approximating the partition function. We need to specify the sample size $S$ in each of the Markov chain simulations needed to compute the ratios.

**Theorem 2.5** *Suppose the sample size $S = \lceil 65\epsilon^{-2}l\exp(2/p)\rceil$ and suppose it is possible to sample* exactly *according to the distributions $\pi_{\beta_i}$, for all $i \in [\![l]\!]$, with polynomially bounded time. Then, there exists an FPRAS with $\epsilon$ as the error parameter for computing the partition function.*

**Proof** The proof uses standard techniques described in (Jerrum & Sinclair, 1996). Let $X_i^{(1)}, \cdots, X_i^{(S)}$ be a sequence of $S$ independent copies of the random variable $f_i$ obtained by sampling from the distribution $\pi_{\beta_i}$, and let $\bar{X}_i = S^{-1}\sum_{j=1}^{S} X_i^{(j)}$ be the sample mean. We have $\mathbb{E}\bar{X}_i = \mathbb{E}f_i = \rho_i$, and $\mathrm{Var}\bar{X}_i = S^{-1}\mathrm{Var} f_i$. The final estimator $\rho = Z(\theta|x)^{-1}$ is the random variable $X = \prod_{i=1}^{l} \bar{X}_i$ with $\mathbb{E}X = \prod_{i=1}^{l} \rho_i = \rho$. Now, consider

$$\frac{\mathrm{Var} X}{(\mathbb{E}X)^2} = \prod_{i=1}^{l}\left(1 + \frac{\mathrm{Var}\bar{X}_i}{(\mathbb{E}\bar{X}_i)^2}\right) - 1$$
$$\leq \left(1 + \frac{\exp(\frac{2}{p})}{S}\right)^l - 1$$
$$\leq \exp\left(\frac{l\exp(\frac{2}{p})}{S}\right) - 1$$
$$\leq \epsilon^2/64 \ ,$$

if we choose $S = \lceil 65\epsilon^{-2}l\exp(2/p)\rceil$ (because $\exp(a/65) \leq 1 + a/64$ for $0 \leq a \leq 1$). By applying

Chebyshev's inequality to $X$, we get

$$\Pr[(|X-\rho|) > (\epsilon/4)\rho] \leq \frac{16}{\epsilon^2}\frac{\operatorname{Var}X}{(\mathbb{E}X)^2} \leq \frac{1}{4},$$

and therefore, with probability at least $3/4$, we have

$$\left(1-\frac{\epsilon}{4}\right)\rho \leq X \leq \left(1+\frac{\epsilon}{4}\right)\rho.$$

Thus, with probability at least $3/4$, the partition function $Z(\theta|x) = X^{-1}$ lies within the ratio $(1 \pm \epsilon)$ of $\rho^{-1}$. Polynomial run time immediately follows from the assumption that we can sample exactly according to the distributions $\pi_{\beta_i}$ in polynomial time. $\square$

We have shown how to approximate the partition function under the assumption that there exists an exact sampler[2]. In fact, it suffices to have only an exact uniform sampler. As we will see in Section 3, it is possible to obtain exact samples from distributions of interest other than uniform if there exists an exact uniform sampler.

### 2.3 Approximating the Gradient of the Log Partition Function

The optimisation problem (1) is typically solved using gradient descent methods which involves gradient-vector multiplications. We now describe how to approximate the gradient-vector multiplication with provable guarantees using concentration inequalities. Let $z$ be a vector in $\mathbb{R}^n$ (where $n$ is also the dimension of the feature space $\phi(x,y)$) with bounded $\ell_2$ norm, i.e., $||z|| \leq G$, where $G$ is a constant. The gradient-vector multiplication is given as

$$\langle \nabla_\theta \ln Z(\theta|x), z \rangle = \mathbb{E}_{y \sim p(y|x,\theta)}[\langle \phi(x,y), z \rangle].$$

We use Hoeffding's inequality to bound the deviation of $\langle \nabla_\theta \ln Z(\theta|x), z \rangle$ from its estimate $\langle d(\theta|x), z \rangle$ on a finite sample of size $S$, where

$$d(\theta|x) = \frac{1}{S}\sum_{i=1}^{S}\phi(x,y_i),$$

and the sample is drawn according to $p(y|x,\theta)$.

Note that by Cauchy-Schwarz's inequality, we have $|\langle \phi(x,y_i), z \rangle| \leq RG$, for all $i \in [\![S]\!]$. Applying Hoeffding's inequality, we then obtain the following exponential tail bound:

$$\Pr(|\langle \nabla_\theta \ln Z(\theta|x) - d(\theta|x), z \rangle| \geq \epsilon) \leq 2\exp\left(\frac{-\epsilon^2 S}{2R^2G^2}\right).$$

---

[2]A similar result can be derived by relaxing the exact sampling assumption and is described in Appendix A.

## 3 Sampling Techniques

We now focus on designing sampling algorithms. These algorithms are needed (i) to compute the partition function using the machinery introduced in the previous section, and (ii) to do inference, i.e., predict structures, using the learned model by solving the optimisation problem $\operatorname{argmax}_{y \in \mathcal{Y}} p(y|x,\theta)$ for any $x \in \mathcal{X}$. Sampling algorithms can be used for optimisation using the Metropolis process (Jerrum & Sinclair, 1996) and other methods like simulated annealing for convex optimisation (Kalai & Vempala, 2006) (note that these methods come with provable guarantees and are not heuristics).

The main contribution of this section is a generic, 'meta' approach that can be used to sample structures from distributions of interest given that there exists a uniform sampler. We start with the design of a Markov chain based on Metropolis process (Metropolis et al., 1953) to sample according to exponential family distributions $p(y|x,\theta)$ under the assumption that there exists an exact uniform sampler for $\mathcal{Y}$. Consider the following chain META: If the current state is $y$, then

1. select the next state $z$ uniformly at random, and

2. move to $z$ with probability $\min\left(1, \frac{p(z|x,\theta)}{p(y|x,\theta)}\right)$.

We now analyse the mixing time of this chain using coupling from the past technique (Propp & Wilson, 1996; Huber, 1998). Coupling from the past (CFTP) is a technique to obtain an exact sample from the stationary distribution of a Markov chain. The idea is to simulate Markov chains forward from times in the past, starting in all possible states, as a coupling process. If all the chains coalesce at time 0, then Propp and Wilson (1996) showed that the current sample has the stationary distribution.

To apply CFTP for META, we need to bound the expected number of steps $T$ until all Markov chains are in the same state. For the chain META, this occurs as soon as we update all the states, i.e., if we run all the parallel chains with the same random bits, once they are in the same state, they will remain coalesced. This happens as soon as they all accept an update (to the same state $z$) in the same step. First observe that, using Cauchy-Schwarz and triangle inequalities, we have

$$\forall y, y' \in \mathcal{Y}: |\langle \phi(x,y) - \phi(x,y'), \theta \rangle| \leq 2BR.$$

The probability of an update is given by

$$\min_{y,y'}[1, p(y|x,\theta)/p(y'|x,\theta)] \geq \exp(-2BR).$$

We then have

$$\begin{aligned}\mathbb{E}T \leq &1 \times \exp[-2BR]+ \\ &2 \times (1 - \exp[-2BR]) \times \exp[-2BR]+ \\ &3 \times (1 - \exp[-2BR])^2 \times \exp[-2BR] + \cdots\end{aligned}$$

By using the identity $\sum_{i=0}^{\infty} i \times a^i = a^{-1}/(a^{-1} - 1)^2$ with $a = (1 - \exp[-2BR])$, we get $\mathbb{E}T \leq \exp(2BR)$. We now state the main result of this section.

**Theorem 3.1** *The Markov chain* META *can be used to obtain an exact sample according to the distribution $\pi = p(y|x, \theta)$ with expected running time that satisfies $\mathbb{E}T \leq \exp(2BR)$.*

Note that the running time of this algorithm is random. To ensure that the algorithm terminates with a probability at least $(1 - \delta)$, it is required to run it for an additional factor of $O(\ln(1/\delta))$ time (Huber, 1998). In this way, we can use this algorithm in conjunction with the approximation algorithm for computing the partition function resulting in an FPRAS.

The implication of this result is that we only need to have an exact uniform sampler in order to obtain exact samples from the distributions $\pi_{\beta_i}$, for all $i \in [\![l]\!]$, needed to approximate the partition function (cf. Theorem 2.5). As we will see in the next section, designing an exact uniform sampler is possible for several combinatorial structures that are of importance in machine learning problems.

We end this section with a few remarks on the bound in Theorem 3.1 and its practical implications. At first glance, we may question the usefulness of this bound because the constants $B$ and $R$ appear in the exponent. But note that we can always set $R = 1$ by normalising the features. Also, the bound on $R$ (cf. Appendix C) could be loose in practice as observed recently by Do et al. (2009), and thus the value of $R$ could be way below its upper bound $\sqrt{\ln |\mathcal{Y}|/\lambda}$. We could then employ techniques similar to those described in (Do et al., 2009) to design optimisation strategies that work well in practice. Also, note that the problem is mitigated to a large extent by setting $\lambda \geq \ln |\mathcal{Y}|$ and $R = 1$.

While in this work we focused on designing a 'meta' approach for sampling, we would like to emphasise that it is possible to derive improved mixing time bounds by considering each combinatorial structure individually. For instance, Randall (2003) analysed the mixing time of a Markov chain to sample from the vertices of a hypercube uniformly at random. It is fairly straightforward to extend this chain to sample according to distributions $\pi_{\beta_i}$ and also to analyse its mixing time using the technique of *canonical paths* (Jerrum & Sinclair, 1996).

## 4 Application Settings

We describe three combinatorial structures with their corresponding application settings in machine learning. For each of these structures, we show how to obtain exact samples uniformly at random. Together with the 'meta' approach presented in the previous section, it is then possible to obtain exact samples of these structures from exponential family distributions considered in this work. Therefore, we have all the necessary ingredients to approximate the partition function.

**Vertices of a hypercube:** The set of vertices of a hypercube is used as the output space in multi-label classification problems (see, for example, Elisseeff and Weston (2001)). An exact sample can be obtained uniformly at random by generating a sequence (of length $d$, the number of labels) of bits where each bit is determined by tossing an unbiased coin.

**Permutations:** The set of permutations is used as the output space in label ranking problems (see, for example, Dekel et al. (2003)). An exact sample can be obtained uniformly at random by generating a sequence (of length $d$, the number of labels) of integers where each integer is sampled uniformly from the set $[\![d]\!]$ *without* replacement.

**Subtrees of a tree:** Let $T = (V, E)$ denote a directed, rooted tree with root $r$. Let $T'$ denote a subtree of $T$ rooted at $r$. Sampling such rooted subtrees from a rooted tree finds applications in multi-category hierarchical classification problems as considered by Cesa-Bianchi et al. (2006) and Rousu et al. (2006). We now present a technique to generate exact samples of subtrees uniformly at random. The technique comprises of two steps. First, we show how to count the number of subtrees in a tree. Next, we show how to use this counting procedure to sample subtrees uniformly at random. The second step is accomplished along the lines of a well-known reduction from uniform sampling to exact/approximate counting (Jerrum et al., 1986).

First, we consider the counting problem. Let $v \in V$ be a vertex of $T$ and denote its set of children by $\delta^+(v)$. Let $f(v)$ denote the number of subtrees rooted at $v$. Now, $f$ can be computed by using the following recursion:

$$f(r) = 1 + \prod_{c \in \delta^+(r)} f(c) . \qquad (2)$$

Next, we consider the sampling problem. Note that any subtree can be represented by a $d$-dimensional vector in $\{0, 1\}^d$, where $d = |V|$. A naïve approach to generate samples uniformly at random would be the following: generate a sequence of $d$ bits where each bit is determined by tossing an unbiased coin; accept this sequence if it is a subtree (which can be tested in polynomial time). Clearly, this sample has been generated uniformly at random from the set of all subtrees. Unfortunately, this naïve approach will fail if the number of acceptances (subtrees) form only a small fraction of the total number of sequences which is $2^d$, because the probability that we encounter a subtree may be very small. This problem can be rectified by a reduction from sampling to counting, which we describe in the sequel.

We will use the term *prefix* to denote a subtree $T'$ included by another subtree $T''$, both rooted at $r$. Let $L(T')$ denote the set of leaves of $T'$. We will reuse the term *prefix* to also denote the corresponding bit representation of the induced subtree $T'$. The number of subtrees with $T'$ as *prefix* can be computed using the recursive formula (2) and is given (with a slight abuse of notation) by $f(T') = (\prod_{v \in L(T')} f(v)) - |L(T')|$. Now, we can generate a sequence of $d$ bits where each bit $u$ with a prefix $v$ is determined by tossing an *biased* coin with success probability $f(u)/f(v)$ and is accepted only if it forms a tree with its prefix. The resulting sequence is an exact sample drawn uniformly at random from the set of all subtrees.

## 5 Conclusions and Future Work

The primary focus of this work was to rigorously analyse structured prediction models using tools from MCMC theory. We designed algorithms for approximating the partition function and the gradient of the log partition function with provable guarantees. We also presented a simple Markov chain based on Metropolis process that can be used to sample according to exponential family distributions given that there exists an exact uniform sampler. While in the application settings considered in this work, we were able to design an exact uniform sampler, we note that this may not be feasible in general for all applications. In such cases, we can design a Markov chain to obtain approximate samples from distributions of interest and also bound its mixing time. This is possible using coupling technique. Indeed, we show how to obtain approximate samples given that there exists an exact uniform sampler and the analysis is given in Appendix B. We note that the coupling technique is much more amenable than the coupling from the past technique to the problem of obtaining approximate samples from a non-uniform distribution given only approximate uniform samples.

If we were to solve the optimisation problem (1) using iterative techniques like gradient descent, then we have to run Markov chain simulations for every training example in order to compute gradients in any iteration of the optimisation routine. We therefore argue for using online convex optimisation techniques (Hazan et al., 2007; Shalev-Shwartz et al., 2007) as these would result in fast, scalable algorithms for structured prediction. Furthermore, it would be interesting to analyse the effects of our approximation guarantees on the regret bounds and convergence rates of online algorithms. Further application settings include correlation clustering (Bansal et al., 2004) which corresponds to sampling equivalence relations. This setting has attracted a lot of attention recently in machine learning problems (Finley & Joachims, 2005; Haider et al., 2007).

## A Approximating the Partition Function using Approximate Samples

In Section 2.2, we designed an FPRAS for approximating the partition function under the assumption that there exists an exact sampler. We now consider the case where we only have approximate samples resulting from a truncated Markov chain.

We first begin with some definitions. Let $\Omega$ denote the state space of a Markov chain with stationary distribution $\pi$ and transition probability matrix $P$. The *mixing time* of a Markov chain is a measure of the time taken by the chain to converge to its stationary distribution. It is measured by the *total variation distance* between the distribution at time $t$ and the stationary distribution.

**Definition A.1** *Let $P^t(u,v)$ denote the t-step probability of transition from $u$ to $v$. The total variation distance at time $t$ is*

$$||P^t, \pi||_{tv} = \max_{u \in \Omega} \frac{1}{2} \sum_{v \in \Omega} |P^t(u,v) - \pi(u)| \ .$$

**Definition A.2** *For $\epsilon > 0$, the mixing time $\tau(\epsilon)$ is given by*

$$\tau(\epsilon) = \min\{t : ||P^{t'}, \pi||_{tv} \leq \epsilon, \ \forall \ t' \geq t\} \ .$$

A Markov chain is *rapidly mixing* if the mixing time is bounded by a polynomial in the input and $\ln \epsilon^{-1}$.

We are now ready to state the main result of this section.

**Theorem A.1** *Suppose the sample size $S = \lceil 65\epsilon^{-2} l \exp(2/p) \rceil$ and suppose the simulation length $T_i$ is large enough that the variation distance of the Markov chain from its stationary distribution $\pi_{\beta_i}$ is at most $\epsilon/(5l \exp(2/p))$. Under the assumption that the chain is rapidly mixing, there* exists an FPRAS with $\epsilon$ as the error parameter for computing the partition function.

**Proof** The proof again uses techniques described in (Jerrum & Sinclair, 1996). The bound $\text{Var} f_i/(\mathbb{E} f_i)^2 \leq \exp(2/p)$ (from Proposition 2.3) w.r.t. the random variable $f_i$ will play a central role in the proof. We cannot use this bound per se to prove approximation guarantees for the partition function $Z(\theta|x)$. This is due to the fact that the random variable $f_i$ is defined w.r.t. the distribution $\pi_{\beta_i}$, but our samples are drawn from a distribution $\hat{\pi}_{\beta_i}$ resulting from a truncated Markov chain, whose variation distance satisfies $|\hat{\pi}_{\beta_i} - \pi_{\beta_i}| \leq \epsilon/5l \exp(2/p)$. Therefore, we need to obtain a bound on $\text{Var} \hat{f}_i/(\mathbb{E}\hat{f}_i)^2$, w.r.t. the random variable $\hat{f}_i$ defined analogously to $f_i$ with samples drawn from the distribution $\hat{\pi}_{\beta_i}$. An interesting observation is the fact that Lemma (2.4) still holds for $\hat{\rho}_i$, i.e., $\exp(1/p) - 1 \leq \hat{\rho}_i \leq \exp(-1/p) + 1$, for all integers $p \geq 3$, and using similar analysis that followed Lemma (2.4), we get $\text{Var}\hat{f}_i/(\mathbb{E}\hat{f}_i)^2 \leq \exp(2/p)$, $\forall i \in [\![l]\!]$.

Also, note that $|\hat{\pi}_{\beta_i} - \pi_{\beta_i}| \leq \epsilon/5l \exp(2/p)$ implies $|\hat{\rho}_i - \rho_i| \leq \epsilon/5l \exp(1/p)$ (using the fact that $\exp(-1/p) \leq \rho_i \leq \exp(1/p)$). Therefore,

$$(1 - \frac{\epsilon}{5l})\rho_i \leq \hat{\rho}_i \leq (1 + \frac{\epsilon}{5l})\rho_i \ . \tag{3}$$

Equipped with these results, we are ready to compute the sample size $S$ needed to obtain the desired approximation guarantee in the FPRAS. Let $X_i^{(1)}, \cdots, X_i^{(S)}$ be a sequence of $S$ independent copies of the random variable $\hat{f}_i$ obtained by sampling from the distribution $\hat{\pi}_{\beta_i}$, and let $\bar{X}_i = S^{-1}\sum_{j=1}^{S} X_i^{(j)}$ be the sample mean. We have $\mathbb{E}\bar{X}_i = \mathbb{E}\hat{f}_i = \hat{\rho}_i$, and $\text{Var}\bar{X}_i = S^{-1}\text{Var}\hat{f}_i$. The final estimator $\hat{\rho} = Z(\theta|x)^{-1}$ is the random variable $X = \prod_{i=1}^{l} \bar{X}_i$ with $\mathbb{E}X = \prod_{i=1}^{l} \hat{\rho}_i = \hat{\rho}$. From (3), we have

$$(1 - \frac{\epsilon}{4})\rho \leq \hat{\rho} \leq (1 + \frac{\epsilon}{4})\rho \ . \tag{4}$$

Now, consider

$$\frac{\text{Var} X}{(\mathbb{E}X)^2} = \prod_{i=1}^{l}(1 + \frac{\text{Var}\bar{X}_i}{(\mathbb{E}\bar{X}_i)^2}) - 1$$

$$\leq \left(1 + \frac{\exp(\frac{2}{p})}{S}\right)^l - 1$$

$$\leq \exp\left(\frac{l \exp(\frac{2}{p})}{S}\right) - 1$$

$$\leq \epsilon^2/64 \ ,$$

if we choose $S = \lceil 65\epsilon^{-2}l\exp(2/p)\rceil$ (because $\exp(a/65) \leq 1 + a/64$ for $0 \leq a \leq 1$). By applying Chebyshev's inequality to $X$, we get

$$\Pr[(|X - \hat{\rho}|) > (\epsilon/4)\hat{\rho}] \leq \frac{16}{\epsilon^2}\frac{\text{Var} X}{(\mathbb{E}X)^2} \leq \frac{1}{4} \ ,$$

and therefore, with probability at least $3/4$, we have

$$(1 - \frac{\epsilon}{4})\hat{\rho} \leq X \leq (1 + \frac{\epsilon}{4})\hat{\rho} \ .$$

Combining the above result with (4), we see that with probability at least $3/4$, the partition function $Z(\theta|x) = X^{-1}$ lies within the ratio $(1 \pm \epsilon)$ of $\rho^{-1}$. Polynomial run time follows from the assumption that the Markov chain is rapidly mixing. $\square$

## B Mixing Time Analysis of META using Coupling

We will use the following lemma in our analysis.

**Lemma B.1** *(Aldous, 1983)* **(Coupling lemma)** *Suppose $\mathcal{M}$ is a countable, ergodic Markov chain. Let $(P, Q)$ be a random process (the coupling). Suppose $t : (0, 1] \to \mathbb{N}$ is a function such that $\Pr(P_{t(\epsilon)} \neq Q_{t(\epsilon)}) \leq \epsilon$, for all $\epsilon \in (0, 1]$, uniformly over the choice of initial state $(P_0, Q_0)$. Then the mixing time $\tau(\epsilon)$ of $\mathcal{M}$ is bounded from above by $t(\epsilon)$.*

**Theorem B.2** *The mixing time of* META *is bounded from above as follows:*

$$\lceil (\ln \epsilon^{-1})/\ln(1 - \exp(-2BR))^{-1}\rceil \ .$$

**Proof** Using Cauchy-Schwarz and triangle inequalities, we have

$$\forall y, y' \in \mathcal{Y} : |\langle \phi(x, y) - \phi(x, y'), \theta\rangle| \leq 2BR \ .$$

The probability of an update is

$$\min_{y,y'}[1, p(y|x, \theta)/p(y'|x, \theta)] \geq \exp(-2BR) \ .$$

The probability of not updating for $T$ steps is therefore less than $(1 - \exp(-2BR))^T$. Let

$$t(\epsilon) = \lceil (\ln \epsilon^{-1})/\ln(1 - \exp(-2BR))^{-1}\rceil \ .$$

We now only need to show that $\Pr(P_{t(\epsilon)} \neq Q_{t(\epsilon)}) = \epsilon$. Consider

$$\Pr(P_{t(\epsilon)} \neq Q_{t(\epsilon)})$$
$$\leq (1 - \exp(-2BR))^{(\ln \epsilon)/\ln(1-\exp(-2BR))}$$
$$= \exp[\ln(1 - \exp(-2BR) + \epsilon - 1 + \exp(-2BR))]$$
$$= \epsilon \ .$$

The bound follows immediately from the Coupling Lemma (B.1). $\square$

## C Bounds on the Norm of the Parameter Vector

We derive a useful bound on the norm of the parameter vector $\theta$. Let $F(\theta) = -\ln p(\theta, Y|X)$. Consider the optimisation problem (1) for MAP estimation:

$$\hat{\theta} = \operatorname*{argmin}_{\theta} F(\theta)$$
$$= \operatorname*{argmin}_{\theta} \left[ \lambda \|\theta\|^2 + \frac{1}{m} \sum_{i=1}^{m} [\ln Z(\theta|x_i) - \langle \phi(x_i, y_i), \theta \rangle] \right]$$

**Proposition C.1** *The norm of the optimal parameter vector $\hat{\theta}$ is bounded from above as follows:*

$$\|\hat{\theta}\| \leq \sqrt{\frac{\ln |\mathcal{Y}|}{\lambda}} \ .$$

**Proof** Consider any $(x, y) \in \mathcal{X} \times \mathcal{Y}$. Denote by $\ell(\theta, x, y)$ the loss function, where $\ell(\theta, x, y) = \ln Z(\theta|x) - \langle \phi(x, y), \theta \rangle \geq 0$, and note that $\ell(0, x, y) = \ln |\mathcal{Y}|$. Therefore, the true regularised risk w.r.t. $\hat{\theta}$ and an underlying joint distribution $D$ on $\mathcal{X} \times \mathcal{Y}$ is

$$E_{(x,y) \sim D}[\ell(\hat{\theta}, x, y)] + \lambda \|\hat{\theta}\|_2^2 \leq F(0) = \ln |\mathcal{Y}| \ .$$

This implies that the optimal solution $\hat{\theta}$ of the above optimisation problem lies in the set $\{\theta : \|\theta\| \leq \sqrt{\ln |\mathcal{Y}|/\lambda}\}$.
$\square$